\documentclass[conference]{IEEEtran}
\IEEEoverridecommandlockouts

\usepackage{pbalance}

\usepackage[all]{nowidow}

\usepackage{subcaption}

\usepackage[americanresistors,americaninductors]{circuitikz}
\usepackage{tikz}
\usepackage{siunitx}

\usetikzlibrary{decorations.pathreplacing,calligraphy, decorations.pathmorphing,patterns}
\usetikzlibrary{shapes.geometric, arrows}

\usepackage{amsmath}
\usepackage{graphicx}
\usepackage[colorlinks=true, allcolors=blue]{hyperref}

\usepackage{cite}
\usepackage{amsmath,amssymb,amsfonts}
\usepackage{algorithmic}
\usepackage{graphicx}
\usepackage{textcomp}
\usepackage{xcolor}
\def\BibTeX{{\rm B\kern-.05em{\sc i\kern-.025em b}\kern-.08em
    T\kern-.1667em\lower.7ex\hbox{E}\kern-.125emX}}

\usepackage{cleveref}

\Crefname{equation}{Eq.}{Eqs.}
\Crefname{figure}{Fig.}{Figs.}
\Crefname{tabular}{Tab.}{Tabs.}

\begin{document}

\title{Electric-Field Induction Motor with Discrete Circuit Elements and Improved Rotor Dynamics}

\makeatletter
\newcommand{\linebreakand}{%
  \end{@IEEEauthorhalign}
  \hfill\mbox{}\par
  \mbox{}\hfill\begin{@IEEEauthorhalign}
}
\makeatother

\author{
  \IEEEauthorblockN{Alex S. Miller}
   \IEEEauthorblockA{\textit{Dept. of Aeronautics and Astronautics} \\
    \textit{Massachusetts Institute of Technology}\\
    Cambridge, USA
    }
  \and
  \IEEEauthorblockN{Jeffrey H. Lang}
  \IEEEauthorblockA{\textit{Dept. of Electrical Eng. and Computer Sci.} \\
    \textit{Massachusetts Institute of Technology}\\
    Cambridge, USA}
}
\maketitle

\begin{abstract}
Electric-field-based actuators and machines lead the design space at the nanoscale and in biological systems, while having the potential to be used in favorable macroscale applications which prioritize minimal mass, performance in a vacuum, or performance in high temperatures. Electric-field induction motors consist of a set of stator electrodes which induce opposite charges onto a movable rotor surface. In earlier work, rotor surfaces for electric-field induction motors have been made of single continuum materials, which are limited in possible resistivity, permittivity and circuitry characteristics. To enhance design freedom, we develop a rotor for electric-field-based machines which may be constructed out of discrete circuit components between a finite number of conductive vanes. Magnetic-field induction motors, which are common in machinery, often utilize ``deep bar'' rotors, consisting of conducting bars which decrease the rotor time constant with increasing frequency using the skin depth effect. We propose, analyze, design and demonstrate a macro-scale electric-field induction motor which uses circuitry on the rotor surface to emulate the  ``deep bar'' design, providing improved torque over a wide range of speeds. 
\end{abstract}


\section{Introduction}

Electric-field-based motors and machines remain an underexplored area of motor-development, with potential applications from medical devices to space technology. The performance of electric-field machines scales with working gap surface area, whereas performance of magnetic-field machines scales with volume \cite{melcher_continuum_1981}; due to this tradeoff, electric-field machines provide leading performance on the micro and nano scale \cite{bart_electroquasistatic_1989}\cite{nagle_analysis_2000}. Indeed, motors and actuators in biological systems are all electric-field-based, employing nano-scale gaps. Researchers have been recently able to make electrostatic motors using folded DNA \cite{pumm_dna_2022}. Electrostatic machines also excel in applications that prioritize minimal mass, such as small aerial vehicles \cite{shen_sunlight-powered_2024} and applications that prioritize low material thickness, such origami foldable actuators \cite{miller_macro-scale_2026}. Macro-scale electrostatic machines have also been explored for industrial equipment targeting high efficiency and long working life \cite{ludois_macroscale_2022}. With multiplicative gains from dielectric gap fluids \cite{ge_dielectric_2016}, increases in machine area-to-gap ratio, and optimizations in motor packaging systems, electric-field motors are beginning to approach the performance of magnetic machines in macro-scale industrial applications \cite{ge_high_2019}\cite{ludois_macroscale_2022}. 

Electrostatic motors have been developed since the 1740s \cite{jefimenko_electrostatic_2011}, with numerous topologies and configurations developed. Direct-current electrostatic motors have been implemented using corona discharge between the rotor and stator for charge transport \cite{jefimenko_electrostatic_2011}; these motors are simple and capable of very-high operating voltage, but limited in efficiency and are constrained to a relatively small range of physical scales due to reliance on corona discharge through the gap medium. Similar commutation techniques have also been implemented with brushes \cite{shen_sunlight-powered_2024}. Variable-capacitance machines use synchronous alternating-current electric fields to drive the rotor from lower-capacitance positions to higher capacitance positions. These motors, analogous to magnetic variable-reluctance machines, have seen extensive development \cite{philp_vacuum-insulated_1977}\cite{bollee_electrostatic_1969}\cite{huanga_study_1993}\cite{ge_3d_2016}, but the requirement that the excitation be synchronized to the rotor position complicates controller design and the configuration of the excitation limits the potential across the gap to only half of the excitation. Permanent electret machines, analogous to permanent magnet machines, use a permanently charged electret in the rotor to maintain high electric fields \cite{jefimenko_electrostatic_2011} \cite{gross_electret_1985}. Like a permanent-magnet motor, these may be used with a commutator for simple operation, but are limited in their capability due to the limited strength of possible electret materials \cite{gross_electret_1985}. Separately-excited motors use an electrically-driven rotor and stator, with a synchronous field pattern, to achieve a optimal gap field configurations for a range of speeds \cite{ghule_sensorless_2021}\cite{ludois_macroscale_2022}. Separately-excited designs are appropriate for low-speeds, and have seen successful development for industrial applications; however, this topology requires a slip-ring to transmit power to the rotor, adding significant mechanical complexity, especially at micro scales and high speeds, and requiring synchronous controller schemes \cite{ghule_sensorless_2021}\cite{ludois_macroscale_2022}.

Electric-field-based induction machines, analogous to magnetic-field induction machines, use a moving electric field excitation in the stator, along with charge-relaxation in the rotor to induce an asynchronously traveling electric field in the rotor \cite{melcher_traveling-wave_1966}\cite{melcher_continuum_1981}, as shown in \Cref{fig:induction_motor_diagram}. Electric-field-based induction machines are often known in the literature as electro-quasistatic induction machines, while they are, in fact, electro-quasistatic, due to the charge relaxation on the rotor; this paper, these terms are used interchangeably. The asynchronous nature of induction machines negates the need for a shaft encoder for most operation, and they have high maximum efficiency. Additionally, since they are asynchronous, high-voltage drive electronics may be simplified, running at a fixed frequency for the course of motor operation.

However, previous electro-quasistatic induction machines relied on a precisely-selected material on the rotor to achieve a determined speed-torque behavior; selecting and depositing an appropriate rotor conductor in the ranges needed for good motor performance is challenging. In previous work, rotor conductor selection required navigating a design tradeoff between choosing either high starting torque and high efficiency at the operating point, a significant tradeoff not experienced by separately-excited machines.

\begin{figure}[t!]
    \centering
    \includegraphics[width=\linewidth]{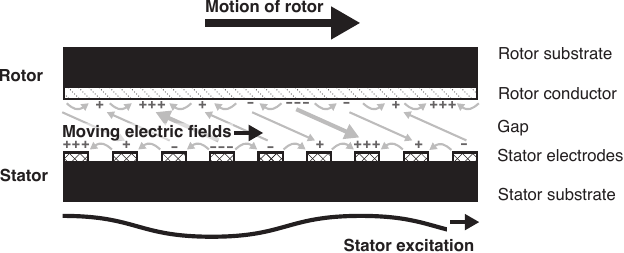}
    \caption{Diagram showing the electromechanical operation of an electric-field induction motor. The stator electrodes are driven with a traveling sinusoidal excitation which induces charge on the rotor surface. Charge relaxation on the rotor surface results in shear force between the rotor and stator, causing the rotor to move.}
\label{fig:induction_motor_diagram}
\end{figure}

Early electric-field induction motors, built in the 1860s, consisted of a cylindrical ebonite rotor surrounded by four copper electrodes, excited with a radial electric field \cite{jefimenko_electrostatic_2011}. These designs were elaborated on with tighter gaps between the rotor and stator, more stator poles, and new rotor insulator materials \cite{bollee_electrostatic_1969}. Subsequent macro-scale induction motors with glass rotors were made to test torque characteristics \cite{dadkhah_experimental_2009} and achieve position control \cite{moser_precise_2002}. With the emergence of micro electro-mechanical manufacturing techniques, many new electric-field-based actuation systems emerged. Induction micromotors were able to achieve speeds in excess of 50,000\,rpm, with gaps on the order of \SI{3}{\micro\meter} \cite{nagle_electric_2005}\cite{livermore_high-power_2004}\cite{steyn_generating_2005}\cite{steyn_self-excited_2009}.

In this paper, we address two of the major challenges that have limited electric-field induction motor designs of the past. First, we implement a rotor conductor based on discrete off-the-shelf components instead of a hard-to-find continuous material coating, facilitating adjustability of design parameters, consistent manufacturing and replicability of results. Second, we design and implement circuits on the rotor of the electro-quasistatic induction motor, which bring new design options for attaining both high starting torque and high efficiency. We design and implement an induction motor testbed, and evaluate the speed-torque characteristics of these new circuit techniques.

\section{Continuum Model} \label{sec:continuum_model}

The electro-quasistatic induction motor in this work uses electric fields which are aligned to the axis of rotation (i.e. an axial-field machine), implemented using printed circuit boards (PCBs) for ease of manufacturing and replication. The techniques discussed in this paper may also be applied to radial-field electro-quasistatic induction machines, built with other processes, with only slight changes to the analysis.

A continuum model of a linear electro-quasistatic motor as well as applicable studies of charge in various physical media was derived by Melcher \cite{melcher_traveling-wave_1966}\cite{melcher_continuum_1981}. Nagle and Krein made further modeling interpretations, adapting the model for use in rotary machines \cite{nagle_electric_2005}\cite{krein_analysis_1995}. The model in this paper follows that of Melcher \cite{melcher_traveling-wave_1966}\cite{melcher_continuum_1981} and Nagle \cite{nagle_analysis_2000}\cite{nagle_electric_2005}, and is summarized in this section. Modifications for discrete resistive elements on the rotor and the rotor circuits are discussed later in this paper.

The charge-relaxation dynamics of the rotor may be characterized by the time-constant, 
\begin{equation}\label{eq:time-constant}
    \tau_{\mathrm{R}} \equiv \epsilon_{\mathrm{eff}}\rho_{\mathrm{eff}} = \frac{\epsilon_{\mathrm{eff}}}{\sigma_\mathrm{eff}},
\end{equation}
where $\epsilon_{\mathrm{eff}}$ is effective permittivity of the motor, and $\sigma_\mathrm{eff}$ is the effective conductivity of the motor \cite{nagle_electric_2005}. These dynamics are a function of the slip frequency 
\begin{equation} \label{eq:slip}
    \omega_{\mathrm{s}} \equiv \omega_{\mathrm{e}}  -  m\omega_{\mathrm{mech}}, 
\end{equation}
which is the difference between the excitation frequency, $\omega_{\mathrm{e}}$, and the mechanical frequency, $\omega_{\mathrm{mech}}$ multiplied by the stator periodicity, $m$. Conceptually, the slip frequency is the electrical frequency experienced by the moving rotor. The stator is excited with a time-varying potential boundary condition processing in the azimuthal direction. Expressions for charge convection and conservation on the rotor boundary, and Gauss's law can be combined with Laplacian field solutions to derive expressions for the electric field in the motor \cite{nagle_electric_2005}. The integrated electromagnetic stress tensor can then be used to solve for torque on the rotor as 
\begin{equation}\label{eq:torque_equation}
    \mathcal{T} = \epsilon_{\mathrm{g}} \pi \hat{V}^2 \int^{R_{\mathrm{o}}}_{R_{\mathrm{i}}} \frac{\tau_{\mathrm{R}} \omega_{\mathrm{s}}}{1+ (\tau_{\mathrm{R}} \omega_{\mathrm{s}})^2} \frac{\epsilon_{\mathrm{g}}}{\epsilon_{\mathrm{eff}}\sinh\left(\frac{mG}{r}\right)} \frac{m^2}{\sinh\left(\frac{mG}{r}\right)} dr
\end{equation}
where $\epsilon_{\mathrm{g}}$ is the gap permeability, $V$ is the amplitude of the stator traveling wave, $R_{\mathrm{o}}$ is the outer radius of the motor electrodes, $R_{\mathrm{i}}$ is the inner radius of the motor electrodes and $G$ is the gap between the rotor and the stator \cite{nagle_electric_2005}. The term $\frac{m}{r} \equiv k \equiv \frac{2 \pi}{\lambda}$, can be considered the spatial wavenumber of the stator traveling wave, with a wavelength of $\lambda$. For a rotor with $\mathcal{N}$ stator phases per wave, the amplitude of the stator traveling wave is $\hat{V} = V_{\mathrm{excitation}}\operatorname{sinc}(1/\mathcal{N})$, where $V_{\mathrm{excitation}}$ is the excitation amplitude of each individual electrode.

\section{Discrete Rotor Resistors} \label{sec:discrete_rotor_elements}

In induction machines, the torque and speed characteristics are shaped by the time constant of the rotor. The rotor time constant, $\tau_{\mathrm{R}}$, is set by the rotor resistance multiplied by the system capacitance \cite{nagle_analysis_2000}\cite{melcher_continuum_1981}.  In previously developed electro-quasistatic induction machines, designers set the time constant by finding appropriate material properties of a homogeneous material. Glass is a common choice for a bulk material \cite{bollee_electrostatic_1969} with well-characterized electromechanics \cite{moser_precise_2002} \cite{moser_electrostatic_2002}. Glass may also be coated with a poorly-conducting material such as Indium-Tin-Oxide (ITO), which decreases the surface resistivity from around \SI{e14}{\ohm} for untreated glass to around \SI{e6}{\ohm} with ITO \cite{jeong_induction_1998}. In addition, other materials have also been used such as moderately doped polysilicon film \cite{nagle_electric_2005}\cite{steyn_generating_2005}, Plexiglas \cite{dadkhah_experimental_2009} and Bakelite \cite{choi_surface-charge_1971}. While deposition techniques do allow for consistent resistivity across the device, it is challenging to find a suitable material for any arbitrary resistivity. To provide more resistivity options, we develop a rotor design around off-the-shelf discrete resistive elements which can match a material of arbitrary resistivity.

\begin{figure}[t!]
    \centering
    \includegraphics[width=\linewidth]{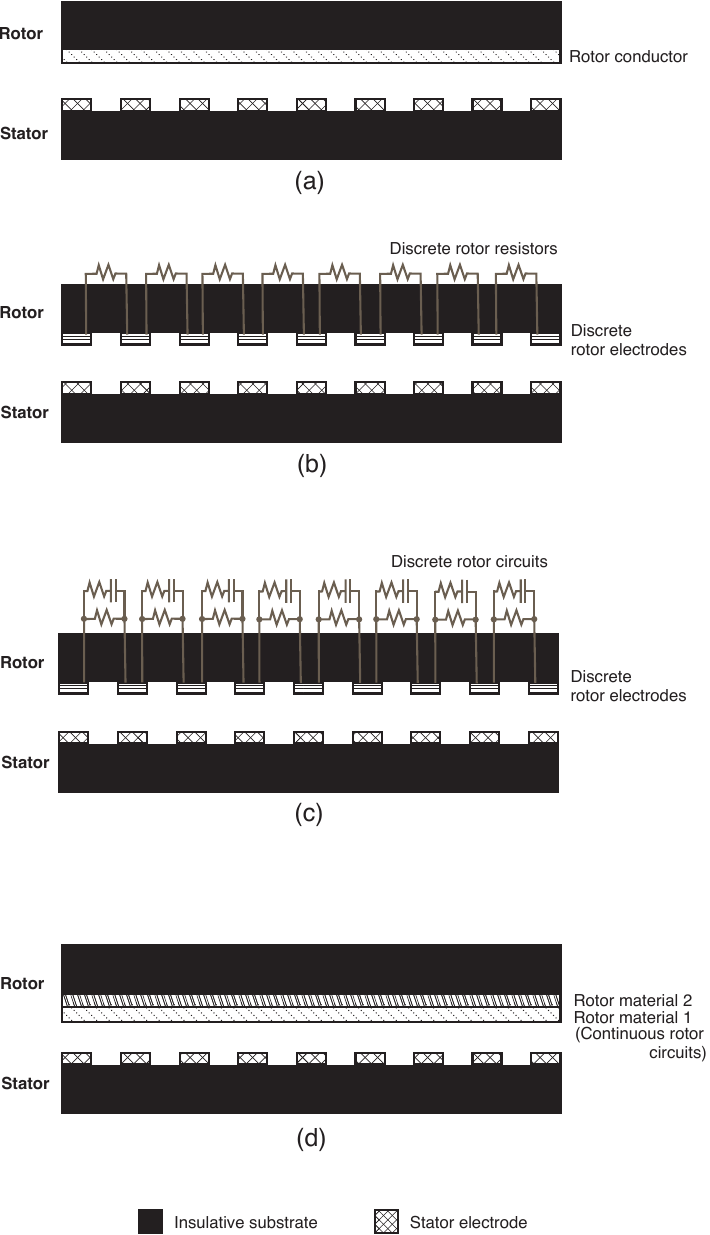}
    \caption{Diagram showing simplified topologies of electro-quasistatic induction motors including (a) the traditional electro-quasistatic induction motor with a continuous rotor conductor material. (b) electro-quasistatic induction motor with discrete rotor resistors such as developed in this work. (c) electro-quasistatic induction motors with circuits on the rotor to shape the speed-torque characteristics, such as developed in this work. (d) electro-quasistatic induction motors with circuits made from two or more layers of continuous material on the rotor with circuits that change the speed-torque characteristics.}
    \label{fig:deep_bar_topologies}
\end{figure}

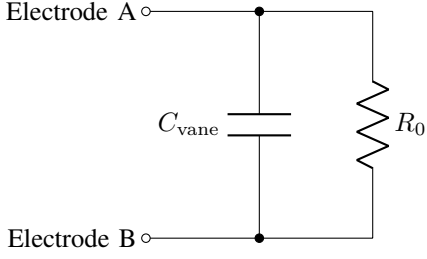
\begin{figure}[!ht]
\centering
\begin{circuitikz}
\sisetup{per-mode=fraction}
\draw 
(4.5,0) to[C, *-*, l={$C_\mathrm{vane}$}]  (4.5,3) 
(3,3) to[short, o-]  (6,3)
(6,3) to[R,l={$R_\mathrm{0}$}]  (6,0) 
(6,0) to[short, -o ]  (3,0)
(3,3) node[left]{Electrode A}
(3,0) node[left]{Electrode B}
;
\end{circuitikz}
\caption{Circuit between rotor vane electrodes on electro-quasistatic induction motor with discrete rotor resistors, $R_\mathrm{0}$, and parasitic vane capacitance, $C_\mathrm{vane}$. }
\label{fig:single_stage}
\end{figure}

To achieve a rotor design with off-the-shelf discrete resistors, we lay out a set of evenly-spaced conductive vanes connected by resistors on a flat printed circuit board substrate. These vanes are spaced out differently than the stator electrodes, such that they intentionally do not match up.

In previous electro-quasistatic induction motor models \cite{melcher_traveling-wave_1966}\cite{melcher_continuum_1981}\cite{nagle_electric_2005}, rotor characteristics are represented using sheet quantities; to fit within those models, we translate lumped parameters to sheet parameters. The sheet resistance quantities in this paper are denoted with the ``$\square$'' symbol. The sheet resistance of a rotor with adjacent vanes connected by lumped resistors of value $R_{\mathrm{0,lumped}}$ is expressed as
\begin{equation}
    R_{\mathrm{0}\square} = R_{\mathrm{0,lumped}} \frac{L_{\mathrm{vane}}}{P_{\mathrm{vane}}}, 
\end{equation}
where $P = \frac{2 \pi R_{\mathrm{avg}}}{\mathcal{K}}$ is the vane pitch where $\mathcal{K}$ is the total number of rotor vanes, $R_{\mathrm{avg}} = (R_{\mathrm{o}} + R_{\mathrm{i}})  / 2$ and $L=R_{\mathrm{o}} - R_{\mathrm{i}}$ is the vane length. The sheet capacitance between two adjacent vane electrodes may be expressed as a function of the lumped capacitance between them as 
\begin{equation}
C_\square = C_{\mathrm{lumped}} \frac{P_{\mathrm{vane}}}{L_{\mathrm{vane}}}.
\end{equation}
The relation between lumped and sheet capacitance has inverse geometric coefficient, since capacitance add in parallel, whereas resistors add in series.

The equivalent sheet resistance from the discrete rotor resistors combined with the resistive properties of the rotor substrate to form the effective motor conductivity, 
\begin{equation}
   \sigma_{\mathrm{eff}} = \frac{m}{r}\frac{1}{R_{0\square}} + \frac{\coth\left(\frac{m \Delta_{\mathrm{ri}}}{r}\right)}{\rho_{\mathrm{ri}}},
\end{equation}
where $\Delta_{\mathrm{ri}}$ is the thickness of the rotor substrate and $\rho_{\mathrm{ri}}$ is the resistivity of the rotor material. While effects of the rotor substrate are negligible for most macro-scale electro-quasistatic induction motors, considering them is important in substrate selection. Total effective permittivity of the motor system can be written as
\begin{equation}
   \epsilon_{\mathrm{eff}} = \frac{m}{r} C_{\mathrm{vane}\square} + \epsilon_{\mathrm{g}} \coth\left(\frac{m}{r}G\right) + \epsilon_{\mathrm{ri}}\coth\left(\frac{m}{r}\Delta_{\mathrm{ri}}\right),
\end{equation}
consisting of the parasitic equivalent sheet capacitance between adjacent rotor vanes, the capacitance of the gap between the rotor and the stator, and the permittivity of the rotor substrate material. To maximize performance, most of $\epsilon_{\mathrm{eff}}$ should be in the gap; $C_{\mathrm{vane}\square}$ and $\epsilon_{\mathrm{ri}}$ should be minimized in actuator design. 

\begin{figure}[t!]
    \centering
    \includegraphics[width=\linewidth]{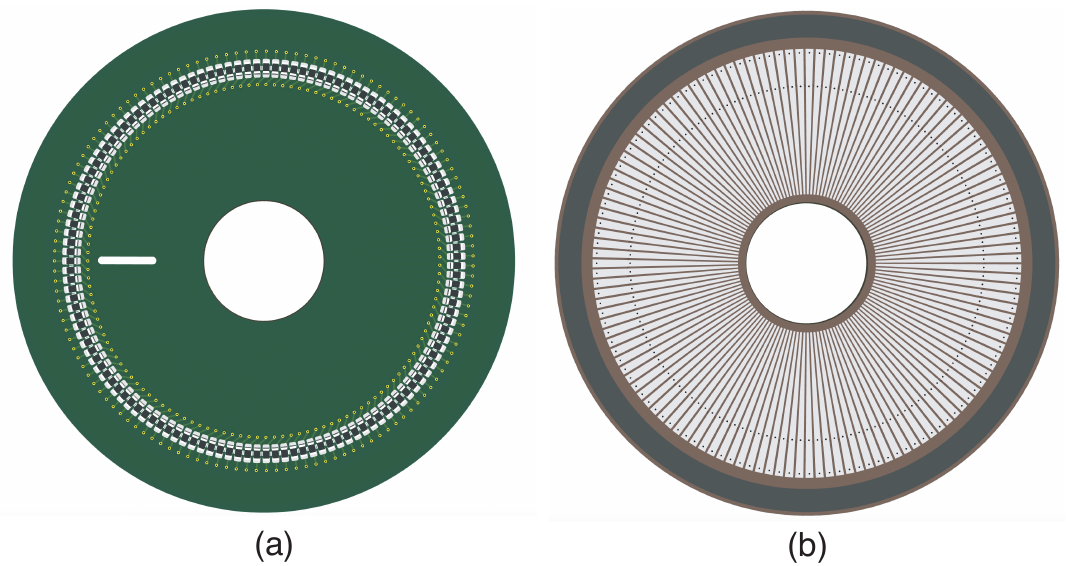}
    \caption{Rotor design for electro-quasistatic induction motor using discrete resistors. (a) Front of rotor circuit board, showing discrete resistors. (b) Back of rotor circuit board, showing electrode vanes.}
    \label{fig:discrete}
\end{figure}

In order to calculate $C_{\mathrm{vane}}$, the vanes may be represented as two half-striplines on the PCB substrate, and modeled accordingly, assuming an infinitely-thick substrate \cite[p.~381]{gupta_microstrip_1996}. The capacitance between two adjacent traces is modeled as
\begin{equation} \label{eq:vane-to-vane}
    C = 2 \epsilon_0(\epsilon_r +1) \frac{K(k_1)}{K'(k_1)} L
\end{equation}
where K(k) and K'(k) are the complete elliptic integral of the first kind, and its complement, respectively, where $K(k)=\int_0^{\pi/2} \frac{1}{\sqrt{1-k^2 \sin^2(\theta)}}d\theta$, and $K'(k_1) = K(k_1')$ where $k_1' = \sqrt{1-k_1^2}$, and where $L = R_\mathrm{o} - R_\mathrm{i}$ is the active length of the rotor vane. The geometric parameter is 
\begin{equation}
k_1 = \frac{S_{\mathrm{re}}}{S_{\mathrm{re}}+ 2W_\mathrm{active}},
\end{equation}
where $S_{\mathrm{re}}$ is the distance between electrodes, and $W_\mathrm{active}$ is the active width of the electrodes \cite[p.~382]{gupta_microstrip_1996}. Since each rotor vane has neighbors on each side, $W_\mathrm{active} = \frac{1}{2}\frac{\mathcal{W}_{\mathrm{R,o}} + \mathcal{W}_{\mathrm{R,i}}}{2}$, the active width is taken as half of the average electrode width, where $\mathcal{W}_{\mathrm{R,i}}$ is the electrode width at the innermost point and $\mathcal{W}_{\mathrm{R,o}}$ is the electrode width at the outermost point.

\section{Rotor Circuits for Improved Rotor Dynamics} \label{sec:deep_bar}

In prior electro-quasistatic induction motor designs, the general shape of the speed-torque curve was fixed, set by $\tau_{\mathrm{R}}$. The drawback of these prior designs was that there was a direct design tradeoff between starting torque and efficiency at operating speed. In prior designs, if the fixed $\tau_{\mathrm{R}}$ was small, there would be high starting torque but low maximum efficiency, and if the fixed $\tau_{\mathrm{R}}$ were large, there would be low starting torque but higher maximum efficiency. Where slip is defined as $s = \omega_{\mathrm{s}} / \omega_{\mathrm{e}}$, $s \in [0,1]$; efficiency is highest when slip is minimized, as $\eta \propto 1-s$. Thus, to design a rotor with both high starting torque and high efficiency, the rotor should have a variable $\tau_{\mathrm{R}}(\omega_{\mathrm{s}})$ which satisfies
\begin{equation} \label{eq:slip_design_inequality}
\lim\limits_{\omega_{\mathrm{s}} \to 0} \tau_{\mathrm{R}}(\omega_{\mathrm{s}}) >  \lim\limits_{\omega_{\mathrm{s}} \to \infty} \tau_{\mathrm{R}}(\omega_{\mathrm{s}}). 
\end{equation}
Since many electro-quasistatic induction motors are designed for high speed, low torque applications \cite{livermore_high-power_2004}, reaching the nominal operating speed while being able to overcome bearing friction at startup is a motivating factor for changing the design landscape.

In conventional magnetic-field induction machines, magnetic fields are induced into a conducting material which dissipates with a frequency-independent time-constant \cite{fitzgerald_electric_2005}\cite{alger_nature_1951}. That is, the time constant of the magnetic field in the rotor is the same at high operating frequencies as it is at low operating frequencies. To increase starting torque, magnetic field induction machines often employ a technique called ``deep bar'' rotors, where the rotor is made out of conducting bars of thickness that varies with distance from the stator. In a good conductor, $\delta \approx 1 / \sqrt{\pi f \mu \sigma}$ where $\delta$ is the skin depth that the wave travels in the medium, $f$ is the frequency of the wave, $\mu$ is the magnetic permeability, and $\sigma$ is the conductivity of the material \cite{fitzgerald_electric_2005}. Since electromagnetic waves go deeper into the conductive material at lower frequencies, this principle can be used to construct frequency-dependent behavior in a rotor \cite{fitzgerald_electric_2005}. In a magnetic induction motor, the thickness and material of the bars can be varied to increase starting torque, while achieving high efficiency at nominal operating speed.

While the construction of ``deep bar'' rotors is a commonly accepted technique in magnetic field induction machines, it has never been used in electro-quasistatic induction machines. In an electro-quasistatic induction machine, the time constant is set by the resistivity of the rotor, and the sum of system capacitance (such as between the different electrode elements), such as in \Cref{eq:time-constant}. The slip frequency, $\omega_{\mathrm{s}}$,  defined in \Cref{eq:slip}, will be highest when the mechanical speed is lowest. To make an electro-quasistatic version of the ``deep bar'' rotor, the time constant of the rotor, $\tau_{\mathrm{R}}$ must vary as a function of slip frequency $\omega_{\mathrm{s}}$, with longer time constants at lower slip frequencies, and shorter time constants at higher slip frequencies. 

For driving modern magnetic induction motors, variable frequency drives (VFDs) provide a method of operating the motor at the optimal slip frequency. However, for electric-field actuators, making high-voltage variable frequency drives with several kilohertz of frequency range is challenging, and requires speed sensing to achieve optimal performance. Thus, fixed-frequency drives in conjunction with the rotor circuitry in this paper offers some of the same design flexibility as variable frequency drives. 

To achieve slip-frequency-dependent rotor time constant, we design and manufacture circuits on top of the rotor, which connect adjacent rotor electrodes. This circuit is shown in \Cref{fig:single_stage}, and consists of adding an additional resistor and capacitor to the rotor design developed in \Cref{sec:discrete_rotor_elements}.

\begin{figure}[!ht]

\centering

\begin{circuitikz}
\sisetup{per-mode=fraction}

\draw 

(4.5,0) to[C, *-*, l={$C_\mathrm{vane}$}]  (4.5,3) 

(3,3) to[short, o-]  (6,3)
(6,3) to[R,l={$R_\mathrm{0}$}]  (6,0) 
(6,0) to[short, -o ]  (3,0)

(6,3) to[C, l={$C_\mathrm{1}$},*-]  (8.5,3)
(8.5,3) to[R,l={$R_\mathrm{1}$}]  (8.5,0) 
(8.5,0) to[short, -* ]  (6,0)

(3,3) node[left]{Electrode A}
(3,0) node[left]{Electrode B}
;
\end{circuitikz}

\caption{Circuit between rotor vane electrodes to achieve a frequency-dependent rotor time constant, $\tau_{\mathrm{R}}$, as a function of slip frequency. This design emulates a ``deep bar'' rotor common in magnetic-field machines, with the goal of achieving both high starting torque and high efficiency at nominal operating speed.}
\label{fig:single_stage}
\end{figure}
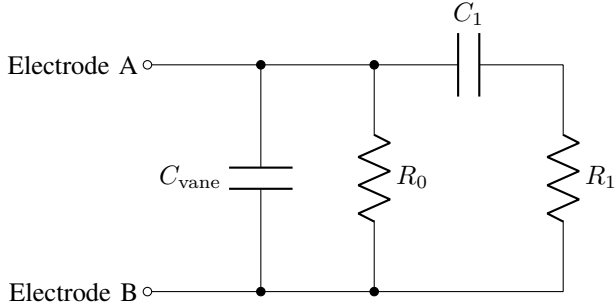

The complex impedance of each rotor circuit may be modeled as 
\begin{equation} \label{eq:impedance}
    \hat{Z}_{\mathrm{deepbar}} = \dfrac{1}{j\omega_{\mathrm{s}} C_{\mathrm{vane}} + \dfrac{1}{R_0} + \dfrac{1}{R_1 +\dfrac{1}{jC_1\omega_{\mathrm{s}}}}}.
\end{equation}
where $R_0$, $R_1$, and $C_1$ are the circuit components shown between neighboring electrodes in \Cref{fig:single_stage}, where $R_0 > R_1$ $C_{\mathrm{vane}}$ is the vane-to-vane parasitic capacitance on the rotor, also shown in \Cref{fig:single_stage}, and $j=\sqrt{-1}$. When $\omega_{\mathrm{s} \to 0}$, the rotor behaves with the character of $R_0$; that is, 
\begin{equation}
    \lim\limits_{\omega_{\mathrm{s}} \to 0}  \hat{Z}_{\mathrm{deepbar}}(\omega_{\mathrm{s}}) = R_0,
\end{equation}
providing high efficiency at low slip values. At high values of $\omega_{\mathrm{s}}$, with the limit taken when $\omega_{\mathrm{s} \to \infty}$ and $C_{\mathrm{vane}} \to 0$,
\begin{equation}
    \lim\limits_{\substack{ \omega_{\mathrm{s}} \to \infty \\ C_{\mathrm{vane}} \to 0}}  \hat{Z}_{\mathrm{deepbar}}(\omega_{\mathrm{s}}) = \frac{R_0 R_1}{R_0+R_1} = R_0 - \frac{R_0^2}{R_0+R_1},
\end{equation}
meaning the rotor has lower impedance, providing improved starting torque, satisfying the design criteria in \Cref{eq:slip_design_inequality}. This frequency-dependent character is incorporated into the model through a new $\sigma_{\mathrm{eff}}$ and $\epsilon_{\mathrm{eff}}$. The rotor conductivity includes the real components of the rotor impedance, such that 
\begin{equation}
   \sigma_{\mathrm{eff}} = \frac{m}{r}\left[\frac{1}{R_{0\square}} + \frac{\omega^2C_{1\square}^2R_{1\square}}{1+(\omega C_{1\square} R_{1\square})^2}\right] + \frac{\coth\left(\frac{m \Delta_{\mathrm{ri}}}{r}\right)}{\rho_{\mathrm{ri}}},
\end{equation}
where $\frac{1}{R_{0\square}} + \frac{\omega^2C_{1\square}^2R_{1\square}}{1+(\omega C_{1\square} R_{1\square})^2} = \operatorname{Re(\hat{Y})}$, $\hat{Y} = 1/\hat{Z}$. The rotor permittivity includes the imaginary (capacitive) components of the rotor impedance, such that 
\begin{multline}
   \epsilon_{\mathrm{eff}} = \frac{m}{r} \left[C_{\mathrm{vane}\square}  + \frac{C_{1\square}}{1+(\omega C_{1\square} R_{1\square})^2} \right] + \\
   \epsilon_{\mathrm{g}} \coth\left(\frac{m}{r}G\right) + \epsilon_{\mathrm{ri}}\coth\left(\frac{m}{r}\Delta_{\mathrm{ri}}\right),
\end{multline}
where $C_{\mathrm{vane}\square}  + \frac{C_{1\square}}{1+(\omega C_{1\square} R_{1\square})^2} = \operatorname{Im(\hat{Y})}$. An implementation a rotor using these principles is shown in \Cref{fig:deep-bar-circuit-board}, which uses networks of discrete resistor and capacitors between vanes.

\begin{figure}[htb!]
    \centering
    \includegraphics[width=\linewidth]{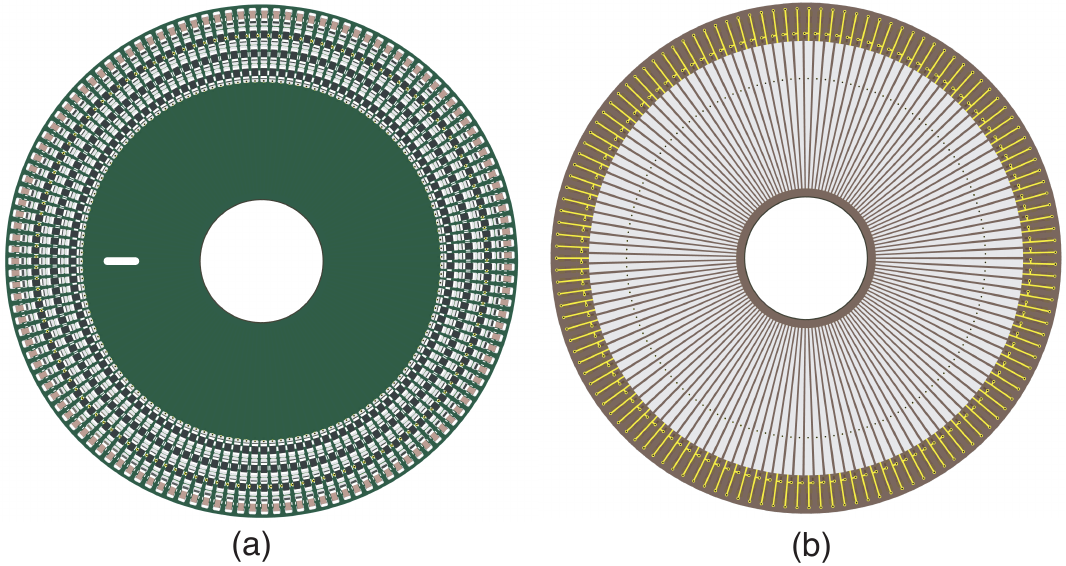}
    \caption{Circuit board design for electric-field induction motor rotor exhibiting improved rotor dynamics. (a) Front of circuit board, showing discrete resistor and capacitor networks. (b) Back of circuit board, showing rotor electrode vanes.}
    \label{fig:deep-bar-circuit-board}
\end{figure}

\section{Motor design} 
\subsection{Electromechanics Design}

The motor in this paper was implemented in an axial field configuration, using PCBs for both the stator and the rotor. The electromechanical parameters of the motor in this paper are listed in \Cref{tab:parameters}. To minimize torque ripple, the stator is designed to have $m \mathcal{N}=132$ electrodes, and the rotor is designed to have $\mathcal{K}=131$ electrodes, such that the electrodes would purposefully not align in any motor position. For rotors with single discrete resistors per vane, resistors values are chosen such that the time constant is shorter than that of the substrate, set by $\rho{_\mathrm{ri}}$ and $\epsilon{_\mathrm{ri}}$, and longer than that set by the parasitic $C_\mathrm{vane}$. Nominal component values for rotors in this paper are listed in \Cref{tab:rotor-parameters}, with Rotor \#5 used for modeling only. The \SI{940}{\mega\ohm} resistances are attained with two (Vishay CRCW0603470MJPTAHR) \SI{470}{\mega\ohm} $\pm$ \SI{5}{\percent} thick film resistors in series, and the \SI{300}{\mega\ohm} resistances are attained with two (Vishay CRCW0603150MJPEAHR) \SI{150}{\mega\ohm} $\pm$\SI{5}{\percent} thick film resistors in series. This series approach was taken to increase consistency and to lower component cost. Capacitors were Murata C0G \SI{1}{\micro\farad}\,$\pm$\,\SI{0.25}{\micro\farad} (GCM1885C2A1R0CA16D), and TDK C0G \SI{2}{\micro\farad}\,$\pm$\,\SI{0.25}{\micro\farad} (C1608C0G1H020CT000N). 

The parasitic capacitance, $C_\mathrm{vane}$ is estimated and noted in \Cref{tab:rotor-parameters} using \Cref{eq:vane-to-vane} with the geometric parameters in \Cref{tab:parameters}. A graph of rotor admittance as a function of $\omega_\mathrm{s}$, calculated with \Cref{eq:impedance} for the rotors in \Cref{tab:rotor-parameters} is shown in \Cref{fig:impedance_plot}. A diagram of the motor test stand implementation is shown in \Cref{fig:test_bench}. 

\begin{table}[h]
    \centering
    \caption{Parameters for motor model. Values for $G$, $R_0$, $R_1$, $C_1$, and $C_{\mathrm{vane}}$ vary, and are described elsewhere.}
    \begin{tabular}{c|c|l}
        Parameter & Value & Description \\
        \hline
        $V_{\mathrm{excitation}}$ & \SI{420}{\volt} & Excitation amplitude per phase   \\
        $R_{\mathrm{o}}$ & \SI{28}{\milli\meter} & Motor outer active radius  \\
        $R_{\mathrm{i}}$ & \SI{9}{\milli\meter} & Motor inner active radius   \\
        $R_{\mathrm{R,o,mech}}$ & \SI{33}{\milli\meter} & Rotor mechanical outer radius   \\
        $R_{\mathrm{R,i,mech}}$ & \SI{8}{\milli\meter} & Rotor mechanical inner radius   \\
        $\Delta_{\mathrm{ri}}$ & \SI{1.6}{\milli\meter}  & Thickness of rotor insulator  \\
        $S_{\mathrm{re}}$ & \SI{0.27}{\milli\meter} & Gap between adjacent rotor electrodes   \\
        $\mathcal{W}_{\mathrm{R,o}}$ & \SI{1.05}{\milli\meter} & Rotor electrode outer width   \\
        $\mathcal{W}_{\mathrm{R,i}}$ & \SI{0.25}{\milli\meter} & Rotor electrode inner width   \\
        $\rho_{\mathrm{ri}}$ & \SI{e12}{\ohm\meter} & Rotor substrate resistivity  \\
        $\epsilon_{\mathrm{g}}$ & 1.0\,$\epsilon_{\mathrm{0}}$ & Gap permittivity  \\
        $\epsilon_{\mathrm{ri}}$ & 4.0\,$\epsilon_{\mathrm{0}}$ & Rotor insulator permittivity   \\
        $m$ & 22 & Stator periodicity  \\
        $\mathcal{N}$ & 6 & Phases   \\
        $\mathcal{K}$ & 131 & Rotor electrodes   \\
        $M_{\mathrm{R}}$ & \SI{11.5}{\gram} & Rotor mass   \\
    \end{tabular}
    \label{tab:parameters}
\end{table}

\begin{table}[h]
    \centering
    \caption{Rotor parameters}
    \begin{tabular}{c|c|c|c|c}
        Rotor \# & $R_0$ & $R_1$ & $C_1$ & $C_\mathrm{vane}$ \\
        \hline
        1 & \SI{300}{\mega\ohm} &  N/A & N/A & \SI{0.68}{\pico\farad} \\
        2 & \SI{940}{\mega\ohm} &  N/A & N/A & \SI{0.68}{\pico\farad} \\
        3 & \SI{940}{\mega\ohm} &  \SI{470}{\mega\ohm} & \SI{1}{\pico\farad} & \SI{0.68}{\pico\farad} \\
        4 & \SI{940}{\mega\ohm} &  \SI{470}{\mega\ohm} & \SI{2}{\pico\farad} & \SI{0.68}{\pico\farad} \\
        5 & \SI{940}{\mega\ohm} &  \SI{470}{\mega\ohm} & \SI{0.3}{\pico\farad} & \SI{0.68}{\pico\farad} \\
    \end{tabular}
    \label{tab:rotor-parameters}
\end{table}

\begin{figure}[ht!]
    \centering
    \begin{subfigure}{0.5\textwidth}
      \includegraphics[width=\linewidth]{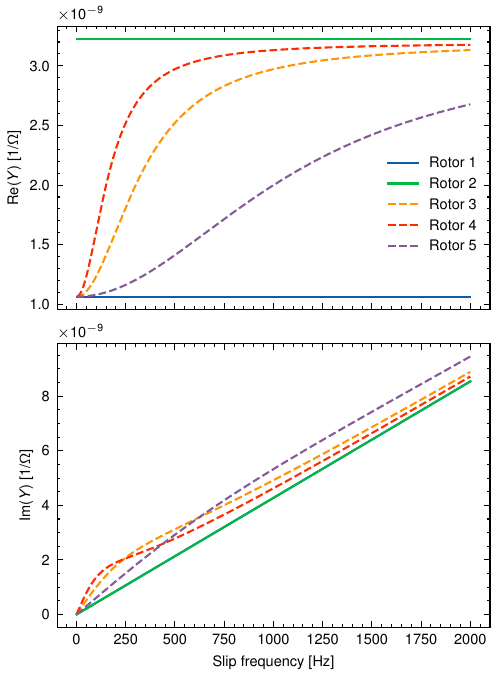}  
    \end{subfigure}
    \caption{Rotor admittance as a function of frequency for the rotors tested in this paper with parameters in \Cref{tab:rotor-parameters}.}
    \label{fig:impedance_plot}
\end{figure}

\begin{figure*}[h!]
    \centering
        \begin{subfigure}[b]{0.3\textwidth}
            \phantomsubcaption  
            \label{fig:test_bench_top_view}
        \end{subfigure}%
        \begin{subfigure}[b]{0.3\textwidth}
            \phantomsubcaption
            \label{fig:test_bench_diagonal}
        \end{subfigure}%
        \begin{subfigure}[b]{0.3\textwidth}
            \phantomsubcaption
            \label{fig:test_bench_side_view}
        \end{subfigure}%

    \begin{tikzpicture}
    \node[text width = \linewidth*0.98] at (0,0) {\begin{center}\includegraphics[width=\linewidth]{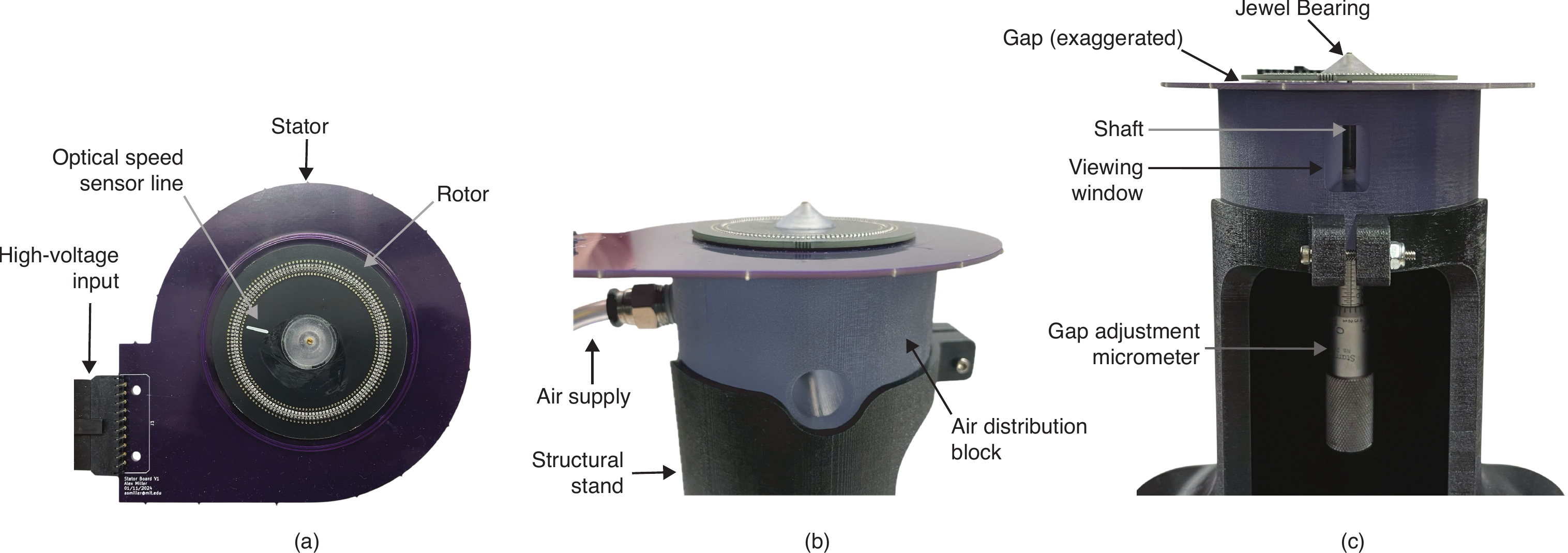}  \end{center}};
    \end{tikzpicture}
    \caption{Test stand design for electro-quasistatic induction motor with discrete components. (a) Top view showing rotor, stator, and contrast patch for optical speed sensor. (b) Slanted view showing hybrid V-jewel and aerostatic bearing with $<$\SI{0.5}{\micro\newton\meter} of static friction. (c) Side view showing adjustable z-axis stage which changes gap thickness, achieving $<$\SI{200}{\micro\meter} gap and accommodating interchangeable test rotors.}
    \label{fig:test_bench}
\end{figure*}

\subsection{Mechanical design}

The motor test setup in this work was designed to accommodate easy rotor interchangeability along with repeatably adjustable gap sizes to demonstrate the electromechanical principles proposed in this work. An axial field disc topology was chosen to facilitate use of a flat stator and rotor made out of easy-to-procure FR4 printed circuit board. Designing bearings that maintain an appropriate gap between the rotor and stator, while constraining radial motion has been a challenge for electrostatic actuators, with previous designs using rigid bearings \cite{bollee_electrostatic_1969}\cite{ludois_macroscale_2022}, aerostatic bearings \cite{moser_electrostatic_2002} \cite{dadkhah_experimental_2009}\cite{livermore_high-power_2004}\cite{steyn_self-excited_2009}, aerodynamic bearings \cite{huanga_study_1993}, diamagnetic levitation \cite{xu_passive_2019}, spacer beads \cite{hosobata_electrostatic_2012}, or electrostatic levitation. The motor in this work operates with a hybrid bearing that includes a sapphire V-jewel (SwissJewel VS-40) riding on a pointed shaft (SwissJewel ST-4D), along with an aerostatic bearing with air flowing through holes on the stator circuit board. The shaft is constrained by a reamed bore in the air distribution block; the bottom end of the shaft sits on the working end of a micrometer head (Starrett No. 263), enabling shaft height adjustment, and thus gap adjustment. The air distribution block was 3D printed in clear resin on a Stratasys J5 printer, and serves to hold the shaft, micrometer head,  and air-supply port, and is bonded to the stator with epoxy. The air distribution block also has a viewport for verifying that the bottom point of the shaft is resting properly on the end of the micrometer head. The entire assembly is mounted on a structural stand, which is clamped to a workbench during testing. This test setup is capable of attaining gap adjustability of $\pm$\,\SI{10}{\micro\meter}.

\subsection{Drive and measurement electronics}

The stator excitation traveling wave is created by running a two-channel signal generator (Rigol DG1022) with \SI{3.3}{\volt}pp sine waves with a \SI{120}{\degree} phase offset into a secondary board which uses op-amps to produce all 6 phases of motor signal. The unamplified signal drives a set of 6 high-voltage linear amplifiers (Apex PA95), supplied by a two high-voltage DC-DC power converters (EMCO F10CT). The amplifier PCB features 6 channels, each with a $\pm$\SI{450}{\volt}pp capability, with voltage regulation provided by a ladder of zener diodes. The amplifier PCB connects with the stator board with the phases A+, A-, B+, B-, C+, and C-, via a shrouded connector (Molex Micro-Fit 3.0).

Speed measurements in this work are taken with a custom laser speed sensor, consisting of a laser module (\SI{5}{\milli\watt}, \SI{650}{\nano\meter}) and a photodiode, connected to a microcontroller (Teensy 4.1) on a printed circuit board. The speed sensor tracks transitions between dark and light on the rotor seen in \Cref{fig:test_bench_top_view}, which occur once per revolution, and streams timestamped tick and speed data to a computer over USB during data collection.

\begin{figure}[h!]
    \centering
    \begin{tikzpicture}
    \node[text width = \linewidth*0.98] at (0,0) {\begin{center}\includegraphics[width=\linewidth]{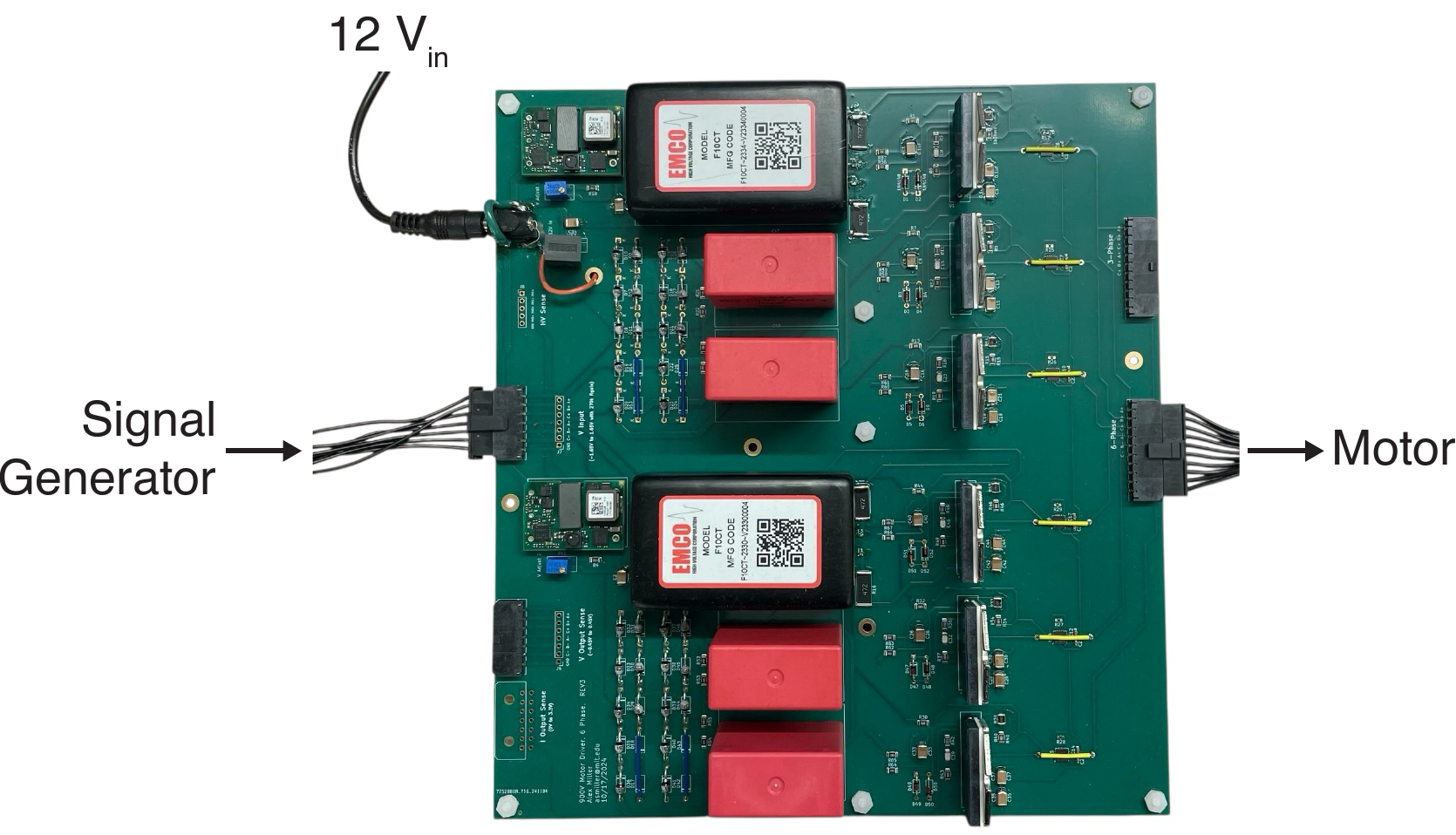}  \end{center}};
    \end{tikzpicture}
    \caption{High voltage amplifier setup for electrostatic motors,  with 6 channels of \SI{900}{\volt}pp linear amplifiers.}
    \label{fig:test_electronics}
\end{figure}

\section{Results}

The electro-quasistatic induction motor in this work was tested with each different rotor design in \Cref{tab:rotor-parameters}. The motor was excited with a range of frequencies $f_\mathrm{e} = \frac{\omega_\mathrm{e}}{2\pi} = $ 10\,--\,\SI{3000}{\hertz}. At each tested frequency, the motor settled at a steady-state speed, which was recorded. The torque relationship in the motor system is
\begin{equation} \label{eq:total_torque_balance}
    \mathcal{T}_\mathrm{output} = \mathcal{T}_\mathrm{motor} - \mathcal{T}_\mathrm{load}.
\end{equation}
where $\mathcal{T}_\mathrm{output}$ is the mechanical output torque, $\mathcal{T}_\mathrm{motor}$ is the electromechanical torque caused by the electro-quasistatic induction motor, and $\mathcal{T}_\mathrm{load}$ is the load torque due to friction and air resistance. In this testing, when the motor is running at a steady state speed, $\mathcal{T}_\mathrm{output} = 0$, so $\mathcal{T}_\mathrm{motor} = \mathcal{T}_\mathrm{load}$. In order to determine the load torque on the motor, power-off spin-down tests are conducted. During these tests, the excitation is turned off such that, $\mathcal{T}_\mathrm{motor} = 0$ and $\mathcal{T}_\mathrm{output} =  -\mathcal{T}_\mathrm{load}$. The output torque may be measured as
\begin{equation}
    \mathcal{T}_\mathrm{output}   = I_{\mathrm{rotor}}\frac{\mathrm{d}\omega_{\mathrm{mech}}}{\mathrm{d} t},
\end{equation}
where $I_{\mathrm{rotor}}$ is the moment of inertia. Thus, we calculate the the motor torque by characterizing load torque and the steady state speed. Multiple trials of spin-down data are shown in \Cref{fig:spindown}; spin-down results are time-invariant and similar across multiple trials.

\begin{figure}[t!]
    \centering
    \includegraphics[width=\linewidth]{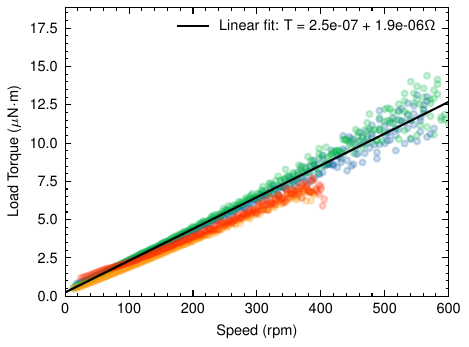}
    \caption{Load characterization of induction motor using power-off spindown tests.}
    \label{fig:spindown}
\end{figure}

The test stand design for this motor uses a micrometer head for relative gap adjustment between the stator and rotor. However, the absolute value of the air gap is needed in order to compare motor performance with torque models. To calculate the absolute value of the air gap, steady-state torque measurements for a wide range of relative gap values, for a fixed excitation frequency. It was verified that the load torque is not strongly dependent on the size of the gap, likely because the gap is sufficiently large that gap-dependent viscous effects are avoided. The minimum gap attainable by the test stand in this paper is set by the air-bearing; the air bearing must have sufficient flow-rate to prevent pull-in electrostatic instability before the rotor has started spinning. Additionally, if the micrometer is set to an infeasible gap given the air-bearing flow, then the rotor will lift off of the V-shaft, losing contact with the radial bearing surface and losing precision of gap adjustability. Indeed, it is possible to attain smaller gap values by starting the motor at a larger gap, then reducing the gap, along with the air supply, during stable rotation. Values of speed and torque are plotted in \Cref{fig:finding_gap} for a fixed excitation of \SI{300}{\hertz}, along with models of speed-state speed from combining the expression for torque in \Cref{eq:torque_equation}, with the point in which the load line intersects the speed-torque curve at steady state per \Cref{eq:total_torque_balance}. The rest of the data in this study were taken with a gap of \SI{300}{\micro\meter}.

\begin{figure}[t!]
    \centering
    \includegraphics[width=\linewidth]{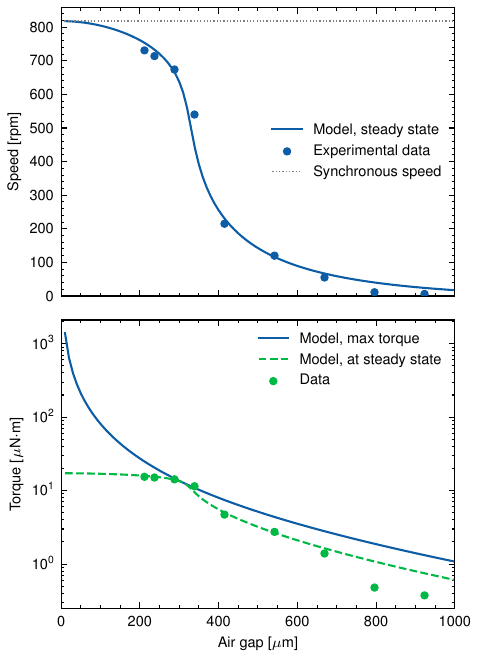}
    \caption{Speed measurements taken for a variety of relative air gap distances between the rotor and the stator, with corresponding torque values using \Cref{eq:total_torque_balance}, for a rotor with \SI{0.94}{\giga\ohm} resistors at a fixed excitation of \SI{300}{\hertz}. The absolute gap values may be calculated using the motor model in \Cref{eq:torque_equation} and \Cref{eq:total_torque_balance}, as represented on the plot. }
    \label{fig:finding_gap}
\end{figure}

The motor in this work is the first to use discrete resistors to set the time constant of an electro-quasistatic induction machine. We evaluate motor torque as a function of slip frequency, $\omega_\mathrm{s}$ by varying the drive frequency, measuring the steady-state speed with the optical speed sensor, then converting it to a steady state torque using the load relation in \Cref{eq:total_torque_balance}. We plot torque experimental results for a rotor with \SI{0.94}{\giga\ohm} resistors in \Cref{fig:discrete_vs_continuous}, which shows the characteristic slip-torque shape of the induction motor, with a peak torque of \SI{13}{\micro\newton\meter}. These data match the model for an electro-quasistatic induction motor with discrete components developed in \Cref{sec:discrete_rotor_elements}. The parasitic capacitance $C_\mathrm{par}$ between the rotor vanes  decreases the torque of the motor; in \Cref{fig:discrete_vs_continuous}, we plot the same model with $C_\mathrm{par} \to 0$, which signifies the torque from an equivalent motor with a continuous rotor conductor material. While the added capacitance of the rotor vanes decreases performance, the benefits of manufacturing repeatability with off-the-shelf parts, and the possibility of making circuits on the rotor make discrete rotor components appropriate for some designs.

\begin{figure}[t!]
    \centering
    \includegraphics[width=\linewidth]{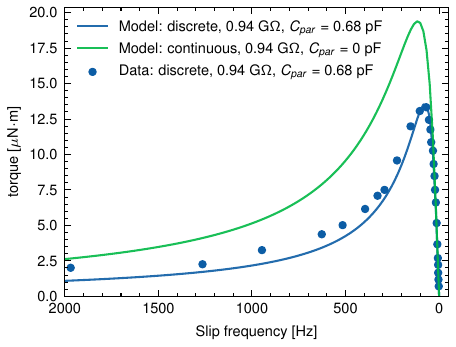}
    \caption{Torque as a function of slip frequency, $\omega_\mathrm{s}$, for a rotor with discrete \SI{0.94}{\giga\ohm} rotor resistors. Experimental data are given and compared to a model which includes parasitic capacitance between rotor vanes. The limit without parasitic capacitance, where $C_\mathrm{par} \to 0$, represents the continuous rotor conductor case. }
    \label{fig:discrete_vs_continuous}
\end{figure}

The motor in this work is also the first electro-quasistatic induction machine to use circuits on the rotor to set a slip-frequency-varying rotor time constant, $\tau_{\mathrm{R}}(\omega_\mathrm{s})$. The rotors in \Cref{tab:rotor-parameters} were all tested, with results shown in \Cref{fig:deep_bar_macro_plot}. The data from the \SI{300}{\mega\ohm} and \SI{0.94}{\mega\ohm} rotors show experimentally how rotor time constant increases with increasing rotor resistance, matching the model in \Cref{eq:torque_equation} with modifications for discrete components in \Cref{sec:discrete_rotor_elements}. Additionally, we demonstrate two rotor models with frequency-dependent rotor time constants. Shown in \Cref{fig:deep_bar_macro_plot}, rotors \#3 and \#4 exhibit the same high-slip and low-slip limits, tending towards the behavior of rotors \#1 and \#2 respectively. The validation of the electro-quasistatic ``deep bar'' concept and model allows for design of new rotors, such as \#5 (model only), which produces a substantial increase in starting torque for a minimal decrease in peak torque. The frequency-dependent rotor designs, mimicking the magnetic ``deep bar'' topology achieve a blend of high efficiency and high starting torque, with a minor decrease to peak torque. Indeed, compared to a single-resistor design of the same efficiency, the rotors with the ``deep bar'' circuit achieve twice the starting torque. This opens up an additional design parameter, allowing electro-quasistatic induction motor designers to pick the most appropriate rotor characteristics for the application. 

\begin{figure}[t!]
    \centering
    \begin{subfigure}[]{0.49\textwidth}
        \centering
        \includegraphics[width=\linewidth]{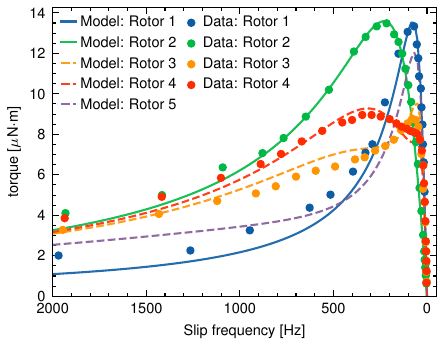}
        \caption{}
    \end{subfigure}

    \begin{subfigure}[]{0.49\textwidth}
        \centering
        \includegraphics[width=\linewidth]{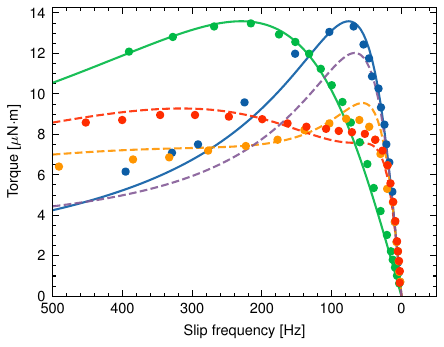}
        \caption{}
    \end{subfigure}%
    \caption{Steady-state torque plotted as a function of slip frequency, plotted with the model from \Cref{sec:continuum_model}, \ref{sec:discrete_rotor_elements} and \ref{sec:deep_bar}, for a (a) wide (b) narrow frequency range, with a shared legend. Circuits on the rotor enable a balance of high starting torque and high efficiency with a minor decrease to peak torque, adding an additional parameter for electrostatic motor designers. }
    \label{fig:deep_bar_macro_plot}
\end{figure}

\section{Conclusions and Discussion}

The motor demonstrates a new electro-quasistatic induction motor which uses discrete rotor resistors to achieve precise and repeatable variations in the rotor time constant. Additionally, the motor in this work is the first demonstration of using circuitry on the rotor to achieve frequency-dependent and speed-dependent electromechanical characteristics towards the goal of achieving both high starting torque and high efficiency at nominal operating speed. In some cases, the ``deep bar'' rotor circuit doubles the starting torque without compromising efficiency, with only a minor decrease in peak torque. The rotor circuits in this work re-create the design flexibility afforded to magnetic-field induction machine designers, where the standard NEMA A, B, C, and D motor types \cite{fitzgerald_electric_2005} denote relations between rotor time constant and slip frequency. In this work, the use of off-the-shelf discrete components makes electro-quasistatic induction motor results more replicable, and the validation of discrete resistors and rotor circuitry in this paper gives electrostatic rotor designers additional freedom to set rotor dynamics on an application-specific basis. 

The design principles developed in this work may be applied to a variety of electro-quasistatic induction machines, including axial field machines, radial field machines, and linear machines. The techniques in this paper may be integrated into electrostatic motors that incorporate dielectric fluids \cite{ge_dielectric_2016}\cite{ludois_macroscale_2022}, micro-sized gaps \cite{livermore_high-power_2004}\cite{nagle_electric_2005}, specialized high-voltage drive electronics \cite{jurgiel_high-voltage_2026}\cite{neugebauer_six-phase_2004}, stacked sets of rotors and stators \cite{ludois_macroscale_2022}\cite{bollee_electrostatic_1969}, and/or folding architectures \cite{miller_macro-scale_2026} to achieve designs more competitive with magnetic motors. In future designs, the rotor vane width and corresponding capacitance, $C_\mathrm{vane}$, may be optimized to the operating conditions. When applying the model of ``deep bar'' rotor circuitry to the electro-quasistatic induction micromotor developed by Nagle \cite{nagle_electric_2005}, these rotor circuits have the potential to facilitate both high starting torque and high efficiency at operating speed, as shown in \Cref{fig:deep_bar_nagle}. 

\begin{figure}[t!]
    \centering
    \includegraphics[width=\linewidth]{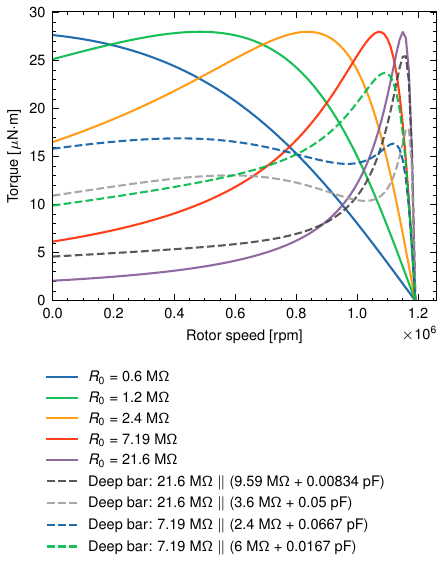}
    \caption{Rotor surface resistance comparison from Nagle \cite{nagle_electric_2005} in the solid lines, with ``deep bar'' options added, shown with the dashed lines.}
    \label{fig:deep_bar_nagle}
\end{figure}

Rotors with circuitry like the ``deep bar'' circuit may also be constructed out of continuum materials, using a resistive material with low permittivity closest to the gap, followed by a less resistive material with a higher permittivity. The rotor circuitry like the ``deep bar'' circuit may, in principle, also be constructed in multiple stages of discrete components instead of one, with a corresponding penalty to the peak torque. The motor in this work used a non-matching number of rotor vanes to the number of stator electrodes to control for torque ripple. Future motors may be designed with rotor vanes that align with the stator electrodes such that a small number of discrete components may be used for the entire rotor, rather than placing components for every rotor vane. 

Electrostatic actuators have a significant competitive advantage in space due to the higher breakdown across gaps under vacuum compared to in air \cite{meek_breakdown_1953}. Combined with the emphasis on lightweight actuators due to launch costs, space applications are a niche suited to electric field machines. Reaction wheels are devices that store and release angular momentum from a spinning inertial disk; these are usually used for rotating satellites and spacecraft. These generally operate at high speeds and low torques which are favorable for electrostatic machines. Current PCB-based satellites are able to point using small magnetorquers which are substantially slower than reaction wheels \cite{gong_design_2022}. Lightweight high speed actuators for reaction wheels would open up mission possibilities with more rigorous pointing requirements to very small satellites, such as those made on a printed circuit board (PCB) \cite{barnhart_low-cost_2009}.

\section{Data and Code Availability}
Data and analysis code for this paper will be made available upon publication.

\section{Acknowledgments}
This material is based upon work supported by the NSF Graduate Research Fellowship under Grant No. 2141064 and by the Fannie and John Hertz Foundation. The authors thank Steven Nagle and the T.J. Rodgers RLE Laboratory for their helpful equipment and advice, Quang Kieu for design advice on the op-amp low-voltage secondary board, as well as Byron Seaborg and David Trumper for helpful conversations.

\IEEEtriggeratref{31}

\bibliographystyle{plainurl}

\end{document}